%% file: main.tex
\documentclass[10pt,twocolumn,letterpaper]{article}

\usepackage{cvpr}              

\usepackage{caption}  
\usepackage{xspace}

\newcommand{\sysname}{ATI\xspace}


\definecolor{cvprblue}{rgb}{0.21,0.49,0.74}
\usepackage[pagebackref,breaklinks,colorlinks,allcolors=cvprblue]{hyperref}

\def\paperID{*****} 
\def\confName{CVPR}
\def\confYear{2025}

\title{ATI: Any Trajectory Instruction for Controllable Video Generation}

\author{Angtian Wang \hspace{0.18in}
Haibin Huang \hspace{0.18in}
Jacob Zhiyuan Fang\hspace{0.18in}
Yiding Yang\hspace{0.2in}
Chongyang Ma
\vspace{4pt}
\\
\textsuperscript{}{ByteDance Intelligent Creation}
\\
\textsuperscript{}{\href{https://anytraj.github.io/}{\texttt{https://anytraj.github.io/}}}
}

\begin{document}
\twocolumn[{%
  \renewcommand\twocolumn[1][]{#1}%
  \maketitle

  \begin{center}
    \includegraphics[width=\textwidth]{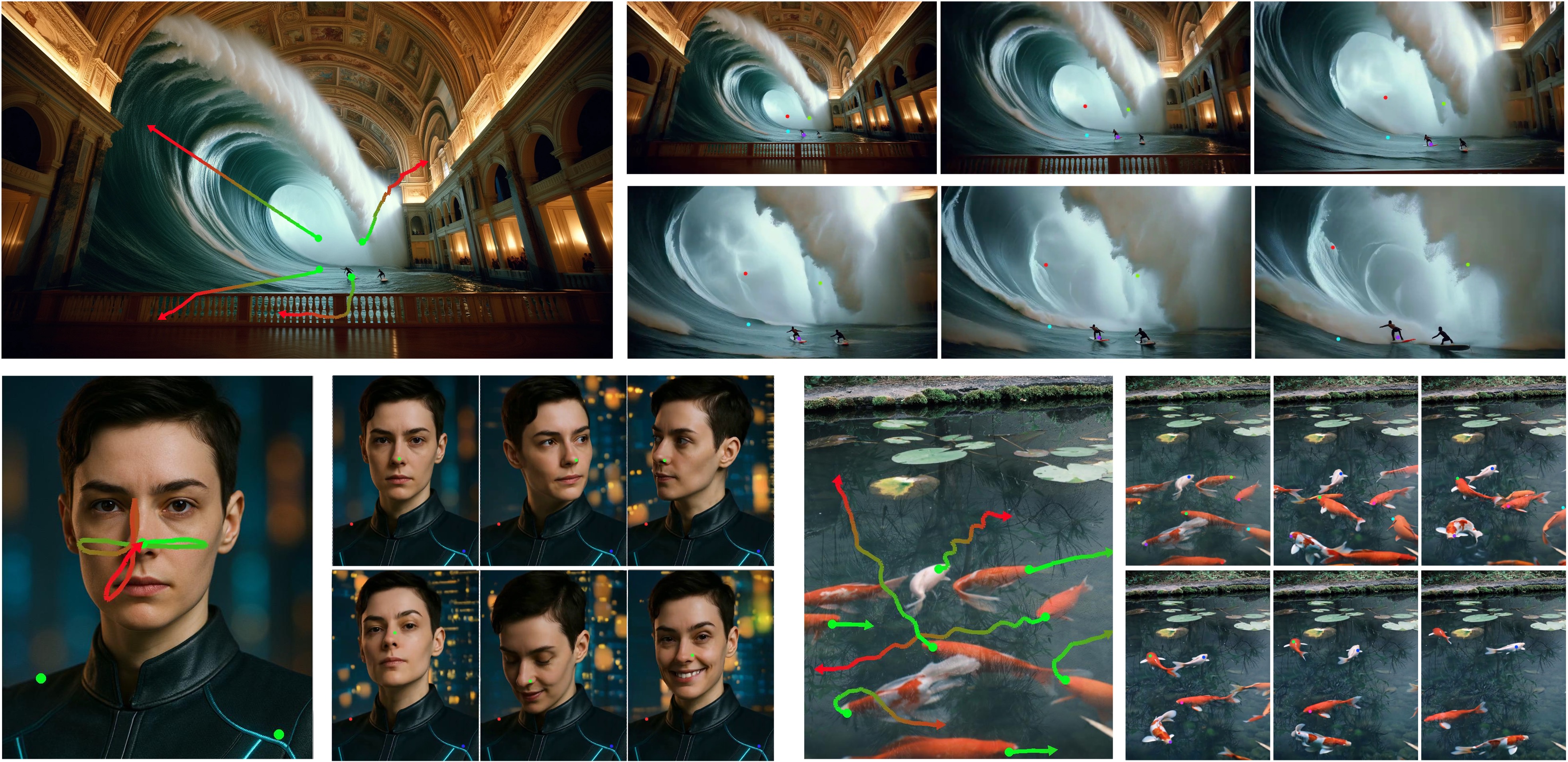}
    \captionof{figure}{%
      \sysname is able to generate a video given an initial frame (left) and a set of user-specified trajectories. Green dots denote the starting points, and red dots indicate the ending points of each trajectory. On the right, we show uniformly sampled frames from the generated video, with colored dots tracking the position of each trajectory point over time.%
    }\label{fig:teaser}
  \end{center}
}]  

\input{sec/0_abstract}    
\input{sec/1_intro}

\input{sec/2_related_work}
\input{sec/3_method}

\input{sec/4_experiment}
\input{sec/5_conclusion}
{
    \small
    \bibliographystyle{ieeenat_fullname}
    \bibliography{main}
}


\end{document}

%% file: sec/0_abstract.tex
\begin{abstract}
We propose a unified framework for motion control in video generation that seamlessly integrates camera movement, object-level translation, and fine-grained local motion using trajectory-based inputs. In contrast to prior methods that address these motion types through separate modules or task-specific designs, our approach offers a cohesive solution by projecting user-defined trajectories into the latent space of pre-trained image-to-video generation models via a lightweight motion injector. Users can specify keypoints and their motion paths to control localized deformations, entire object motion, virtual camera dynamics, or combinations of these. The injected trajectory signals guide the generative process to produce temporally consistent and semantically aligned motion sequences. Our framework demonstrates superior performance across multiple video motion control tasks, including stylized motion effects (e.g., motion brushes), dynamic viewpoint changes, and precise local motion manipulation. Experiments show that our method provides significantly better controllability and visual quality compared to prior approaches and commercial solutions, while remaining broadly compatible with various state-of-the-art video generation backbones.

\end{abstract}

%% file: sec/1_intro.tex
\section{Introduction}

\begin{figure*}[!t]
    \centering
    \includegraphics[width=\linewidth]{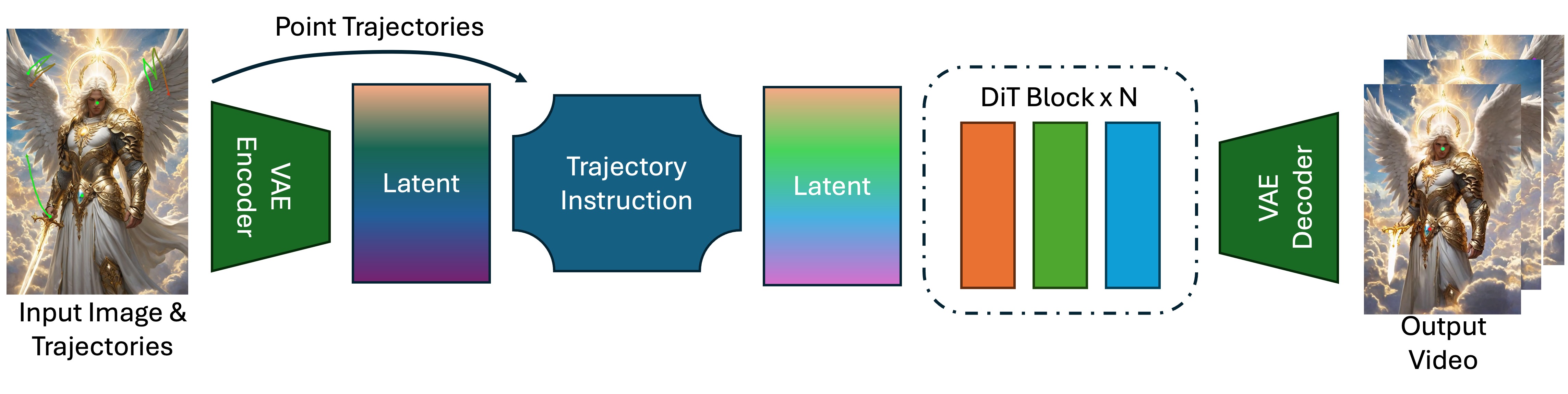}
    \vspace{-8mm}
    \caption{\sysname takes an image and user specified trajectories as inputs. The point-wise trajectories are injected into the latent condition for the generation. Videos are decoded from the latent denoised from the DiT.}
    \label{fig:pipeline}
\end{figure*}

Recent advances in video generation models ~\cite{AlignLatents,hong2022cogvideo,HunyuanVideo,lin2024open,Lumiere,MovieGen,opensora,seawead2025seaweed,Step-Video,SVD,VideoCrafter2,wan2025,yang2024cogvideox,xie2024progressiveautoregressivevideodiffusion, guo2024i2v} have demonstrated remarkable capabilities in synthesizing realistic and diverse video content. However, precise control over motion remains a significant challenge, particularly when users require fine-grained direction over how specific elements move within the generated sequence. Current approaches  ~\cite{niu2024mofa,MotionDirector,MotionBooth,MotionCtrl,Motioni2v} typically address different types of motion control—such as camera movement, object translation, or local deformation—through separate specialized modules, leading to fragmented workflows and inconsistent results.
Motion control in video generation encompasses a spectrum of manipulations, from camera operations (panning, zooming, rotation) to object-level translations and localized movements of specific regions. These motion types are inherently related and often need to be coordinated to achieve desired visual effects. The separation of these controls in existing systems limits creative expression and requires users to navigate multiple interfaces or models.

In this paper, we propose a unified trajectory-based framework that addresses this limitation by treating all forms of motion control through a common lens. Our key insight is that diverse motion effects can be represented as trajectories of specific points within the scene, whether these points are anchored to local features, whole objects, or used to indicate camera perspective changes. By defining motion as trajectory paths for user-selected key points, we establish a consistent, intuitive interface for motion specification.
Our approach builds upon state-of-the-art image-to-video generation models ~\cite{wan2025,seawead2025seaweed}, augmenting them with a specialized motion injector module. This module processes trajectory information and projects it into the latent space of the pre-trained video generation model, effectively guiding the synthesis process to follow the specified motion paths. Importantly, our method does not require retraining the base video model, making it adaptable to different generation architectures.
We demonstrate the versatility of our framework through extensive experiments across various motion control tasks. Our results show that the unified approach not only simplifies the user workflow but also produces higher quality motion than previous methods that handle different motion types separately. The framework excels in challenging scenarios that require coordination between camera movement and object motion, outperforming both specialized academic methods and commercial video generation products in both control precision and visual quality.
The main contributions of our work include:
\begin{itemize}
    \item A unified framework for motion control in video generation that seamlessly integrates camera movements, object-level motion, and local deformations through trajectory-based guidance.
    \item A motion injector module that effectively projects user-specified trajectory controls into the latent space of pre-trained video generation models.
    \item Comprehensive evaluation demonstrating superior performance across various motion control tasks compared to previous methods and commercial products.
    \item Demonstration of compatibility with different base video generation models, highlighting the approach's flexibility and broad applicability.
\end{itemize}
\label{sec:intro}

%% file: sec/2_related_work.tex
\section{Related Work}

Motion-controlled video generation aims to synthesize temporally coherent videos with user-defined motion guidance, which aims to manipulate the camera motion and object movement. Methods such as CamI2V~\cite{zheng2024cami2v}, CameraCtrl~\cite{he2024cameractrl}, and CamCo~\cite{xu2024camco} encode camera trajectories using Plücker coordinates to achieve fine-grained camera path conditioning. Others like ViewCrafter~\cite{yu2024viewcrafter} and I2VControl-Camera~\cite{feng2024i2vcontrol} leverage 3D scene reconstruction from a single image to generate point cloud renderings that guide camera perspectives during generation. Additionally, there exist some training-free approach like~\cite{hou2024training,yu2024zero}. ~\cite{kuang2024collaborative} proposes collaborative diffusion methods which address consistent multi-view synthesis with controllable cameras.

Another important aspect of Motion-controlled video generation is object motion control. Various strategies are used to guide object trajectories, \eg, optical flow-based methods, such as DragNUWA~\cite{yin2023dragnuwa}, Image Conductor~\cite{li2025image}, DragAnything~\cite{wu2024draganything}, and MotionBridge~\cite{tanveer2024motionbridge}, utilize sparse or dense flow to control object displacement. Others like MOFA-Video~\cite{niu2024mofa} and Motion-I2V~\cite{shi2024motion} directly learn dense motion fields to guide generation. Bounding-box-based control is employed in Boximator~\cite{wang2024boximator} and Direct-a-Video~\cite{yang2024direct}, while methods like LeviTor~\cite{wang2024levitor} incorporate depth and clustering for accurate 3D motion guidance. Newer works like ReVideo~\cite{mou2024revideo}, Peekaboo~\cite{jain2024peekaboo}, and Trailblazer~\cite{ma2024trailblazer} expand this paradigm with interactive or trajectory-aware modules. Particularly, training-free method like ~\cite{yu2024zero} injects motion trajectories by decomposing the task into `out-of-place' and `in-place' motion animation and leverage layout-conditioned image generation for motion generation.

Recent works further advances to motion control that simultaneously handles camera and object motion. MotionCtrl~\cite{MotionCtrl} introduces explicit modules to support concurrent control, while Motion Prompting~\cite{geng2024motion} encodes motion tracks to guide both scene and subject dynamics. Perception-as-Control~\cite{chen2025perception} proposes a 3D-aware representation that fuses motion perception with generation. VidCraft3~\cite{zheng2025vidcraft3} builds unified, disentangled control across multiple motion modalities.

%% file: sec/3_method.tex
\section{Method}

We propose \sysname (Figure \ref{fig:pipeline}), a diffusion-based video generation framework that enables Fine-grained feature-level Instruction of Trajectories. Specifically, \sysname introduces a Gaussian-based motion injector to encode trajectory signals, spanning local, object-level, and camera motion, directly into the latent space of a pretrained image-to-video diffusion model. This enables unified and continuous control over both object and camera dynamics.

\begin{figure*}
    \centering
    \includegraphics[width=\linewidth]{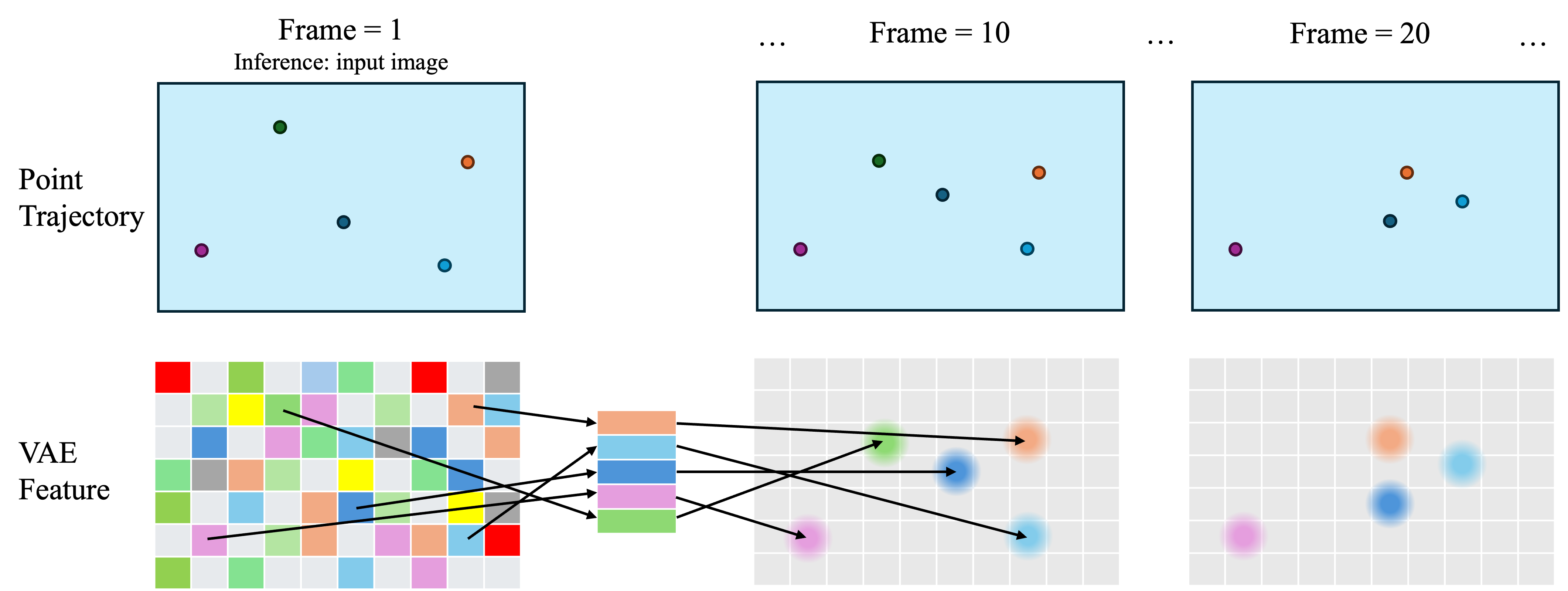}
    \caption{Trajectory Instruction module computes a latent feature from a point's trajectory. During inference, given the point's location in the first frame (\ie, the input image), we sample the feature at that location using bilinear interpolation. We then compute a spatial Gaussian distribution for each visible point on its corresponding location in every subsequent frame.}
    \label{fig:feature}
\end{figure*}

\subsection{Conditional Video Generation Models}
Recent advances in diffusion models have revolutionized generative modeling for both images and videos. In the video domain, these models aim to synthesize realistic, temporally consistent sequences. When extended to conditional generation, the objective is to generate videos based on specific inputs such as text, images, or motion cues, enabling fine-grained control over content, appearance, and motion. Prominent architectures like Diffusion Transformers (DiT) achieve state-of-the-art performance by integrating spatiotemporal modeling with conditional guidance.
Let a video be denoted as $\mathbf{x}_0 \in \mathbb{R}^{T \times H \times W \times C}$, where $T$ is the number of frames. Let $\mathbf{c}$ represent a conditioning signal (e.g., a text prompt, an image, or a trajectory).

The diffusion model defines a forward noising process that progressively corrupts the video with Gaussian noise:
\begin{equation}
q(\mathbf{x}_t | \mathbf{x}_{t-1}) = \mathcal{N}(\mathbf{x}_t; \sqrt{1 - \beta_t} \mathbf{x}_{t-1}, \beta_t \mathbf{I})
\end{equation}
\begin{equation}
q(\mathbf{x}_t | \mathbf{x}_0) = \mathcal{N}(\mathbf{x}_t; \sqrt{\bar{\alpha}_t} \mathbf{x}_0, (1 - \bar{\alpha}_t) \mathbf{I})
\end{equation}
where $\bar{\alpha}_t = \prod_{i=1}^{t} (1 - \beta_i)$.

A neural network $\epsilon_\theta$ learns to reverse the noising process by predicting the added noise, conditioned on $\mathbf{c}$:
\begin{equation}
\mathcal{L}_{\epsilon} =  \left\| \epsilon - \epsilon_\theta(\mathbf{x}_t, t, \mathbf{c}) \right\|^2
\end{equation}

In this report, we choose Seaweed-7B~\cite{seawead2025seaweed} and Wan2.1‑14B~\cite{wan2025} as the base video generation models, where the denoising model $\epsilon_\theta$ is implemented using a DiT architecture ~\cite{Peebles2022DiT}.


\subsection{Gaussian Model for Trajectory Instruction of Feature}
\label{sec:condition}


We propose a Gaussian model for feature‐level instruction of point trajectories. Specifically, for each trajectory point, we assign a weight \(P(f \mid l_{i,j,t})\) to every pixel \((i,j)\) in the latent space.

For a point trajectory \(\phi_t = (x_t, y_t)\) at frame \(t\), we define:
\[
P(f \mid l_{i,j,t})
\;=\;
\exp\!\Bigl(-\tfrac{\|\phi_t - (i,j)\|^2}{2\,\sigma}\Bigr),
\]
where \(\sigma\) is a predefined constant. In practice, we set \(\sigma = \tfrac{1}{440}\) so that the Gaussian weight decays to half its maximum at the nearest diagonal pixel.

As illustrated in Figure~\ref{fig:feature}, we first pass the input image $I$ through the VAE encoder $\Phi$ to obtain a latent feature map
\begin{equation}
L_I = \Phi(I) \in \mathbb{R}^{H\times W\times C}.
\end{equation}
We then extract, for each trajectory point, a $C$‑dimensional feature vector $f$ at its initial position $\phi_0 = (x_0, y_0)$ by bilinearly sampling from $L_I$ whenever $(x_0,y_0)$ does not lie exactly on an integer grid coordinate.  

In Figure~\ref{fig:feature}, the bottom‑left panel depicts the latent feature grid as colored cells; arrows indicate the precise sub‑pixel sampling locations for each trajectory point. The inset in the middle shows how these sampled values assemble into the latent feature vector $f$. Finally, the bottom‑right panels visualize the spatial Gaussian masks—centered at the trajectory locations $\phi_t$ in subsequent frames—computed as
\begin{equation}
P\bigl(f \mid l_{i,j,t}\bigr)
= \exp\ \bigl(-\|\phi_t - (i,j)\|^2 / (2\,\sigma)\bigr),
\end{equation}
which softly distributes the feature $f$ across neighboring latent pixels to guide the image‑to‑video generator with fine‑grained control.

\subsection{Tail Dropout Regularization}
\label{sec:tail_dropout}

In practice, when a user‑specified point trajectory terminates before the end of the video, the model often hallucinates spurious occluders around the final annotated frame. We attribute this to our training labels: any point that falls off its ground‑truth track is marked as “occluded” or “out of frame,” which inadvertently teaches the model to introduce occlusions whenever a trajectory ends.

To mitigate this, we introduce a \emph{Tail Dropout Regularizer}. During training, with probability $p$ (set to $0.2$ in our experiments), we sample a dropout frame
\[
t_d \;\sim\;\mathcal{U}\{0,1,\dots,T\},
\]
where $T$ is the full trajectory length. We then truncate the trajectory via set the visibility of that point to $0$ after frame $t_d$, effectively simulating an early termination. This encourages the model to learn that missing future points do \emph{not} imply occlusion.

Empirically, applying Tail Dropout significantly reduces visual distortions and the appearance of unintended occluders when trajectories end before the final frame at inference time.

\subsection{Data Collection}
\label{sec:data-annotation}
We create our training dataset by first processing 5\,million high–quality video clips,  which are filted to contain no scene cuts and to meet strict aesthetic criteria—and then selecting 2.4\,million clips exhibiting strong object motion. 
To generate point trajectory annotation, we apply TAP-Net~\cite{doersch2022tap} to each selected clip as follows:
\begin{enumerate}
  \item On the first frame, uniformly sample $N = 120$ points such that initial pairwise distances are approximately equal.
  \item Track these seed points throughout the clip using TAP-Net.
  \item Record the trajectory of each point tracker on each frame $t$:
    \begin{itemize}
      \item \textbf{Trajectory} $(x_{t},y_{t})$, the 2D coordinates.
      \item \textbf{Visibility} $v_{t}\in\{0,1\}$, indicating if the point is visible.
    \end{itemize}
\end{enumerate}
All trajectories and visibility flags are stored to support downstream model training and evaluation.
During training, for each video clip, we randomly select 1 to 20 points.

%% file: sec/4_experiment.tex
\section{Experiments}

We integrate \sysname into two different video generation frameworks: Seaweed-7B \cite{seawead2025seaweed} and Wan2.1-14B \cite{wan2025}. In Sec.~\ref{sec:exp:implemenation}, we detail our training and inference setups. We evaluate \sysname on both frameworks, providing qualitative comparisons in Sec.~\ref{sec:exp:qualitative} and quantitative analyses in Sec.~\ref{sec:exp:quantitative}.

\begin{figure*}
    \centering
    \includegraphics[width=\linewidth]{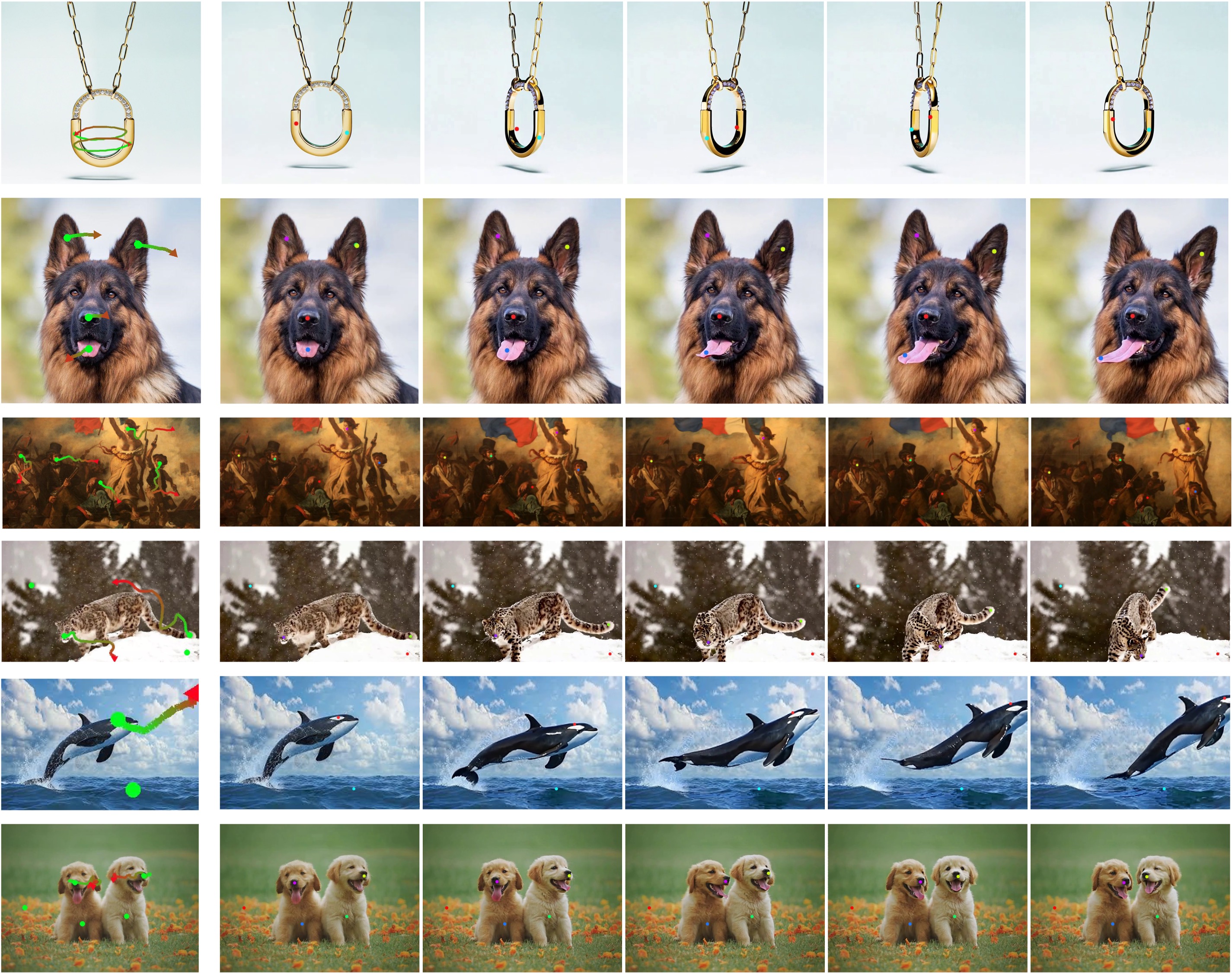}
    \caption{Object Motion Control. Left: the input image overlaid with user‑specified trajectories—green dots mark each trajectory's start point, and arrows mark each end point. Endpoint color encodes trajectory length, indicating that some trajectories span only part of the generated video. Right: five frames uniformly sampled from the generated video. Dot colors serve only to distinguish between trajectories.}
    \label{fig:object}
\end{figure*}

\begin{figure*}
    \centering
    \includegraphics[width=\linewidth]{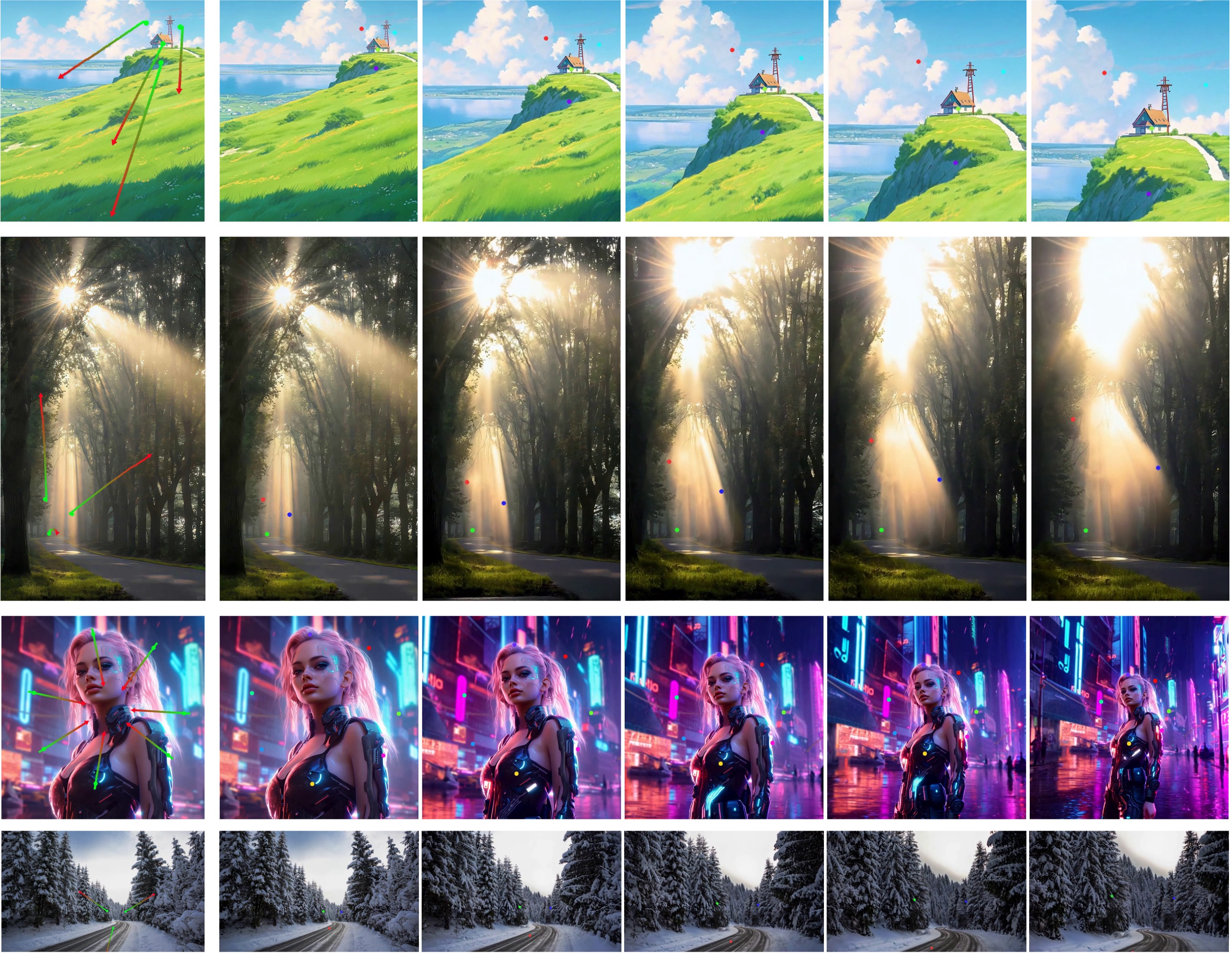}
    \caption{Video generation results with camera control. Left: Input image superimposed with user specified trajectories. Right: Five frames uniformly sampled from the generated video.}
    \label{fig:camera}
\end{figure*}

\begin{figure*}
    \centering
    \includegraphics[width=\linewidth]{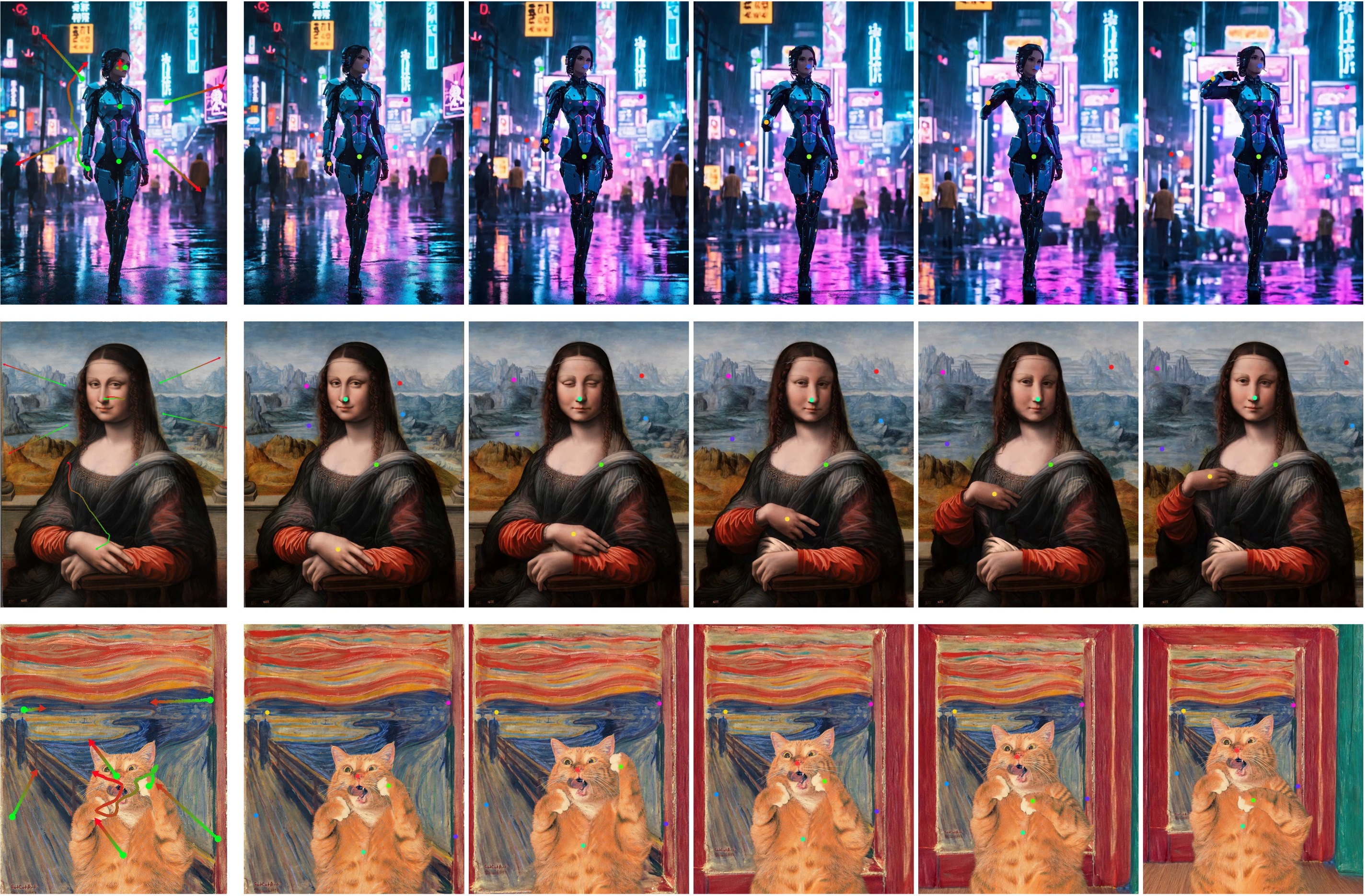}
    \caption{Video generation results with coherent control of camera and object motion. Left: Input image superimposed with user specified trajectories. Right: Five frames uniformly sampled from the generated video.}
    \label{fig:combine}
\end{figure*}

\begin{figure*}
    \centering
    \includegraphics[width=\linewidth]{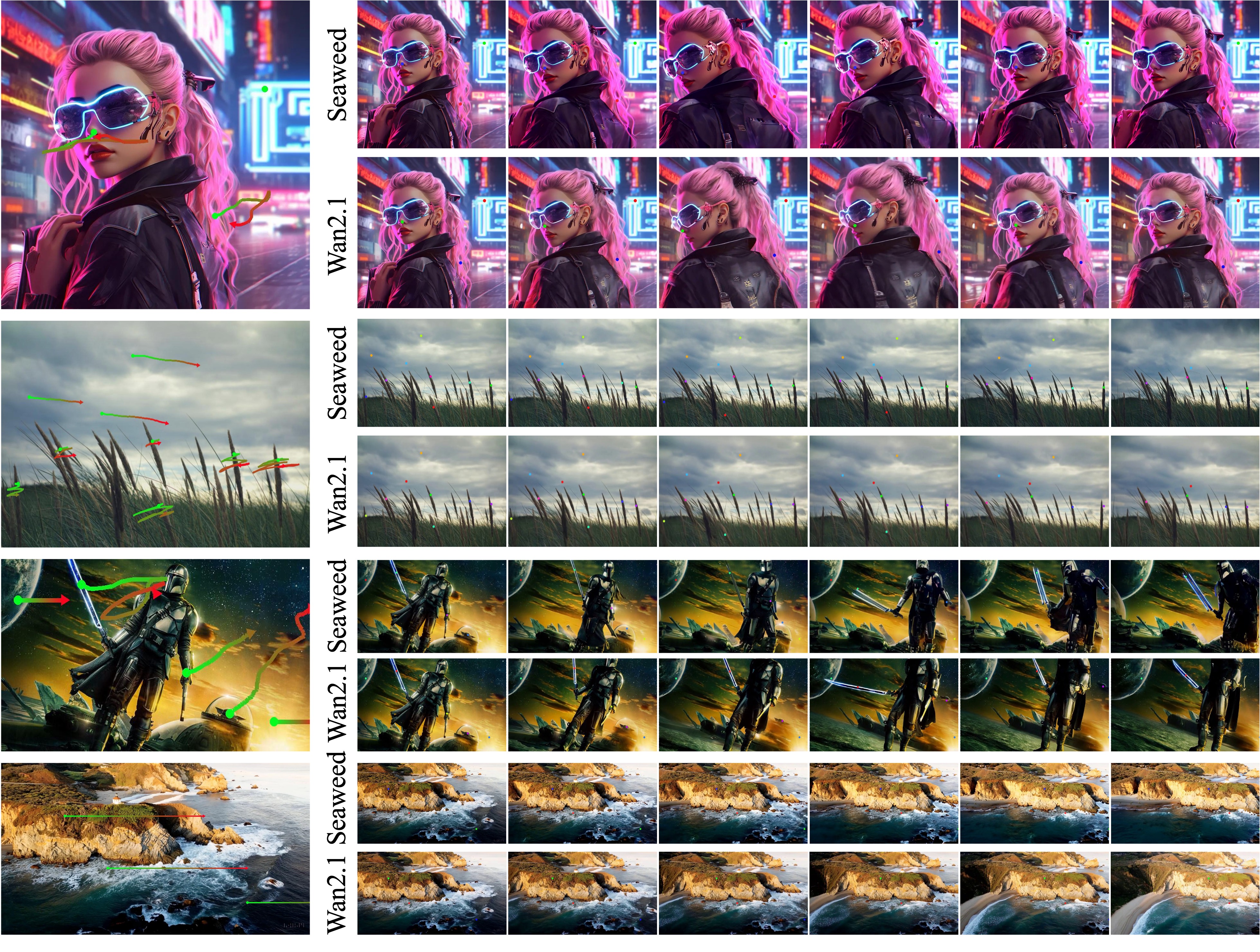}
    \caption{Qualitative comparison for \sysname video generation with different backend models.}
    \label{fig:comparison}
\end{figure*}

\subsection{Implementation Details}
\label{sec:exp:implemenation}

We integrate \sysname into two video generation frameworks: Seaweed-7B \cite{seawead2025seaweed} and Wan2.1-14B \cite{wan2025}. Our implementations build on the pre-trained I2V model by injecting the trajectory instruction between the preprocessing stage and the patchify layer. For both models, we fine-tune all DiT parameters for 50,000 iterations using 64 GPUs with 80 GB of VRAM each. All other training hyperparameters follow the standard I2V fine-tuning setup.

\paragraph{Training and inference time.}
Incorporating the \sysname module into our video generation pipeline does not significantly affect training or inference times. After 15,000 iterations, both models achieve satisfactory trajectory-following performance. During inference, both the Seaweed-7B \sysname and Wan2.1-14B \sysname models generate a five-second, 480p video in approximately 8 GPU-minutes.

\paragraph{Wan2.1 \sysname model details.}
In the Wan2.1 variant, we handle first-frame conditioning by inserting black frames immediately after the initial RGB image and feeding this RGB sequence into the VAE encoder to extract latent features. As a result, the latent stream includes features corresponding to the black frames. We then blend the trajectory instruction features with these black-frame VAE features according to the probability scheme described in Sec.~\ref{sec:condition}.

\paragraph{Interactive trajectory editor.}
We provide an interactive trajectory editor for creating and refining point trajectories on a single input image. The tool allows users to draw and adjust trajectories, place static points to denote stationary objects, and apply global camera motions such as horizontal panning or zooming in and out.

\subsection{Qualitative Results}
\label{sec:exp:qualitative}

We present video generation results from our \sysname model using trajectories created with the tools described in Sec \ref{sec:exp:implemenation}. Unless otherwise stated, all examples use the Seaweed‑7B \sysname model. 

Figure \ref{fig:object} illustrates outputs for trajectories that emphasize \textbf{object motion} and deformation. In the left‑hand insets, we overlay the initial frame with the user‑defined point trajectories: green dots mark each trajectory's start point, and arrows mark its end. The color of each endpoint also encodes trajectory length, since some paths span only part of the generated video. On the right, we show five frames uniformly sampled from the generated video. Dot colors in each frame serve only to distinguish between trajectories. 

Figure \ref{fig:camera} demonstrates the \textbf{camera control} capabilities of our \sysname model. Moving a set of points radially outward from the image center at a constant speed creates a smooth zoom‑in effect. Combining this radial motion with a uniform horizontal translation lets us target the zoom on a specific region. By anchoring a static trajectory on the subject while applying the zoom to the background, we reproduce a classic dolly‑zoom (as the last example shown). However, if all trajectories consist solely of planar horizontal shifts or zooms, the generated video's content may remain static, exhibiting only 2D camera movement.

Figure \ref{fig:combine} illustrates videos generated under \textbf{simultaneous camera and object motion control}. You can create these trajectories by first drawing camera‑movement paths and then editing selected ones for object motion, or by defining object‑motion trajectories first and subsequently applying the camera‑movement transformation.

Figure \ref{fig:comparison} shows a qualitative comparison between the Wan2.1 \sysname model and the Seaweed \sysname model. Overall, we observe that Seaweed \sysname demonstrates slightly better trajectory-instruction-following ability, which may be attributable to differences in how the input latent is zero-conditioned (see Sec. \ref{sec:exp:implemenation}). On the other hand, we observer richer motion for the Wan2.1 \sysname model on those unconstrained locations.

Overall, we observe that \sysname achieves a high success rate in generating videos that follow the user-specified trajectories, except in the following cases:
\begin{itemize}
    \item Very rapid movements (\eg, when a point travels half the image width in two frames), which can prevent the model from accurately following the trajectory.
    \item Trajectories requiring object disassembly (\eg, forcing an object to split into multiple parts), leading to either failure to follow the trajectory or unnatural distortions (\eg, generating an extra cat head).
\end{itemize}

Notably, our approach handles intersecting point trajectories successfully, even when tracking points overlap at certain time steps. We also observe an interesting phenomenon: the \sysname model often finds alternative, realistic solutions to satisfy the user’s trajectory instructions (for example, rotating the camera rather than applying implausible object deformations).

\subsection{Quantitative Results}
\label{sec:exp:quantitative}

\begin{table}[]
    \centering
    \small
    \begin{tabular}{c|ccc}
    \toprule
        \sysname Base Model & \textbf{Acc@0.05} & \textbf{Acc@0.01} & \textbf{App. Rate} \\
         \midrule
        Seaweed-7B \cite{seawead2025seaweed}& 59.0 & 36.0 & 67.9 \\
        \hline
        Wan2.1-14B \cite{wan2025} & 55.9 & 34.7 & 65.5 \\
    \bottomrule
    \end{tabular}
    \caption{Quantitative results for trajectory instruction following ability of \sysname using different base models.}
    \label{tab:score}
\end{table}

As shown in Table \ref{tab:score}, we quantitatively evaluate the ability of \sysname models to follow user‐specified point trajectories. First, we collect 100 image–trajectory pairs; for each image, we manually draw between one and ten point trajectories. We then evaluate all \sysname models on this test set. For each generated video, we use TAP‐Net to track the points from the first‐frame user inputs and compute the per‐frame error distance \(d\) between the TAP‐Net outputs and the ground‐truth trajectories. We introduce three metrics to assess tracking accuracy: \textbf{Acc@0.01}, the percentage of frames where the point distance is less than \(0.01\times\) the image diagonal; \textbf{Acc@0.05}, the percentage of frames where the point distance is less than \(0.05\times\) the image diagonal; and \textbf{Appearance Rate}, the proportion of frames in which the tracker correctly predicts a point as visible whenever the user‐specified trajectory is present. We report the average value of each metric over the entire test set.



%% file: sec/5_conclusion.tex
\section{Conclusion}

In this paper, we introduce \sysname, a unified trajectory‐based control framework that seamlessly integrates camera movement, object translation, and fine‐grained local motion within a single latent‐space injection module.
Our experiments demonstrate that this cohesive approach not only outperforms prior modular methods and commercial systems in both controllability and visual quality, but also remains agnostic to the choice of underlying video generation model. 
In the future, we will further enhance the control capabilities to ensure that object motion better follows both real-world physics and user inputs.